\documentclass[10pt,twocolumn,letterpaper,table]{article}

\usepackage{cvpr}
\usepackage{times}
\usepackage{epsfig}
\usepackage{graphicx}
\usepackage{amsmath}
\usepackage{amssymb}
\usepackage{xcolor}
\usepackage{multirow}
\usepackage{graphicx}
\usepackage{pgfplots}
\usepackage{subfigure}
\usepackage{verbatim}
\usepackage{mathrsfs}
\usepackage{adjustbox}
\usepackage{pgf,tikz}
\usepackage{authblk}
\tikzstyle{startstop} = [rectangle, rounded corners, minimum width=2cm, minimum height=.75cm,text centered, text width=2cm, draw=black, fill=black!20]
\tikzstyle{io} = [trapezium, trapezium left angle=70, trapezium right angle=110, minimum width=2cm, minimum height=.75cm, text centered, draw=black, fill=blue!30]
\tikzstyle{decision} = [diamond, minimum width=2cm, minimum height=.75cm, text centered, draw=black, fill=green!30]
\tikzstyle{process} = [rectangle, minimum width=2cm, minimum height=.75cm, text centered, text width=2cm, draw=black]
\tikzstyle{arrow} = [thick,->,>=stealth]

\usepackage{color} 

\newcommand{\cb}[1]{{\cellcolor{black! #1}$#1$}}
\newcommand{\cw}[1]{{\cellcolor{black! #1}$\color{white} #1$}}
\newcommand*\samethanks[1][\value{footnote}]{\footnotemark[#1]}
\usepackage[pagebackref=true,breaklinks=true,letterpaper=true,colorlinks,bookmarks=false]{hyperref}

\cvprfinalcopy 

\def\cvprPaperID{2332} 

\ifcvprfinal\fi
\begin{document}

\title{Action Recognition with Image Based CNN Features }

\author[1]{Mahdyar Ravanbakhsh\thanks{Authors contributed equally.}}
\author[2]{Hossein Mousavi\samethanks}
\author[3,4]{Mohammad Rastegari}
\author[2]{\\Vittorio Murino}
\author[3]{Larry S. Davis}
\affil[1]{DITEN, University of Genoa, Genova, Italy}
\affil[2]{PAVIS, Istituto Italiano di Tecnologia, Genova, Italy}
\affil[3]{University of Maryland, College Park}
\affil[4]{The Allen Institute for AI}

\maketitle

\begin{abstract}
Most of human actions consist of complex temporal compositions of more simple actions. Action recognition tasks usually relies on complex handcrafted structures as features to represent the human action model. Convolutional Neural Nets (CNN) have shown to be a powerful tool that eliminate the need for designing handcrafted features. Usually, the output of the last layer in CNN (a layer before the classification layer -known as $fc7$) is used as a generic feature for images. In this paper, we show that $fc7$ features, per se, can not get a good performance for the task of action recognition, when the network is trained only on images. We present a feature structure on top of $fc7$ features, which can capture the temporal variation in a video.  To represent the temporal components, which is needed to capture motion information, we introduced a hierarchical structure. The hierarchical model enables to capture sub-actions from a complex action. At the higher levels of the hierarchy, it represents a coarse capture of action sequence and lower levels represent fine action elements. Furthermore, we introduce a method for extracting key-frames using binary coding of each frame in a video, which helps to improve the performance of our hierarchical model. We experimented our method on several action datasets and show that our method achieves superior results compared to other state-of-the-arts methods. 
\end{abstract}

\section{Introduction}
  A current challenge in computer vision is understanding how neural network models, which have been so successful in static image analysis tasks like recognition and segmentation, could be extended to video. 
\begin{figure}
	\begin{center}
		\includegraphics[scale=.31]{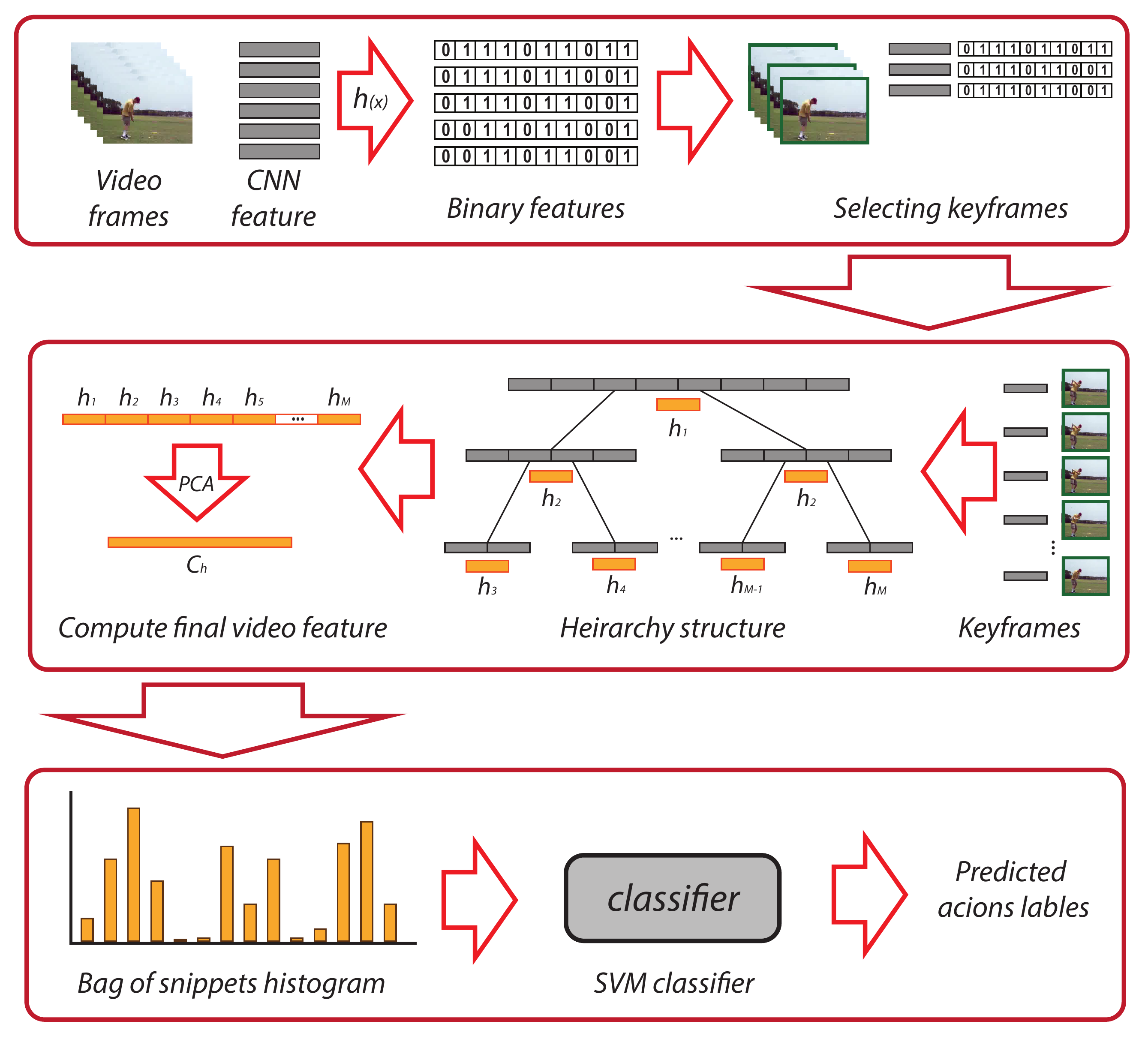}
	\end{center}
	\caption{\textbf{Overview of our method;} given a video, we first compute CNN features for each frame then map them into a short binary code space. We track the changes in each bit of the binary codes across the time to select key-frames. Video snippets between consecutive key-frames feed into a hierarchical decomposition. PCA applies on each level of hierarchy to reduce the dimension in each level. We stack all levels as a vector representation for a video snippet. In the last steps, we build a histogram of temporal words for all the videos and train a classifier to predict action labels.}
	\label{fig:overview}
\end{figure}

 These deep learning models provide a unified approach to image analysis pipeline in a single learning frame work under a deep neural network structure \cite{alexnet, zisserman14a, googlenet}. They take raw images as input and produce a category label for the image as an output. Given a large amount of training data the traditional backpropagation algorithm effectively learns the parameters in the network. A very successful neural network structure for image classification is the Convolutional Neural Network (CNN) \cite{alexnet}. Several CNN architectures have shown state-of-the-arts performance on object recognition \cite{alexnet, zisserman14a, googlenet, overfeat}. The output of the last layer (the layer before the classification layer), known as $fc7$, can be used as a generic image descriptor for other recognition tasks (\eg attribute classification\cite{zhang2014panda}, scene recognition\cite{zhou2014learning}). 

Learning similar networks for videos has remained a challenge. It requires a large amount of supervised video data and very extensive computational resources. There have been efforts on designing CNN for videos \cite{3dcnn21013,zisserman14,wang0T15}. In this paper, we show how $fc7$ features, trained image datasets, can be used as a video descriptor. Training a CNN with images is much cheaper than with videos; therefore, having a video feature constructed  by image based CNN represents a major saving of computational costs. The key-idea behind our method is to track the changes in the $fc7$ feature space across the time. We showed that even a very similar frames may have different $fc7$ features, and that difference captures important properties of video motion. We build a hierarchical structure that captures the differences between frames of a video in a coarse to fine manner. Figure \ref{fig:overview} shows an over view of our approach. In order to focus resources on meaningful sub-videos, we introduce a novel key-frame extraction method using binary coding of frames. Our binary key-frame extraction improves performance on the action recognition task. Our experiments shows state-of-the-arts results in several standard datasets for human action recognition.

 The rest of our paper is organized as follows. In Sect.~\ref{secrelated} reviews the related work on action recognition using both shallow and deep architectures. In Sect.~\ref{secpyramid} introduces CNN Pyramid Architecture for Action Recognition, afterward we describe our experimental cases and the results in Sect.~\ref{secexperiment}. Furthermore, the application of proposed method will be available on \href{https://github.com/matt-rb/cnn_actin_pyramid}{github}.

\subsection{Related work}\label{secrelated}

Action recognition methods can be divided into two categories of approaches: classical models which use hand-crafted features, and deep models using deep-learned features.\\
\noindent\textbf{Classical Models} are standard approach to address the video classification problem using local features~\cite{dollar2005, wang09evaluationof, willems2008}. They have three major stages:
First, extracting local visual features (informative regions of videos) by dense or spare interest points~\cite{hoha, pengWWQ14, pengZQP14}, which tends to track the moving objects in the video. Dense interest points sampling perhaps the most powerful such method~\cite{wang09evaluationof}. Second, the extracted features are encoded in the form of Bag of Features (BoF) to represent the entire video in a vector space. Third, the vectors, along with their category labels, are used to construct a classifier (\eg SVM) to categorize the actions. 

Furthermore, range of 3D (x,y,t) representation method have been developed to extract spatio-temporal descriptors around the detected local interest points, such as Harris3D~\cite{laptev2005space}, Histogram of Gradient (HOG)~\cite{dalal2005}, and Histogram of OpticalFlow (HOF)~\cite{hoha}. However, several works have been done to improve recognition performance with different video representations models~\cite{wangQT14, wang2013, Sadanactionbank}.

 In this paper, we replaced the hand-crafted local descriptors with CNN representation to achieve the high level semantic information.\\
\noindent\textbf{Deep Models} use deep neural networks for action recognition. Deep Neural Networks recently have shown successful results on image-based recognition  tasks~\cite{krizhevsky2012imagenet, zisserman14a, zeilerF13}; also there have been a number of studies have been a deep architectures for action recognition~\cite{zisserman14, karpathy2014large, wang0T15, 3dcnn21013, cheronLS15}. Ji \etal~\cite{3dcnn21013} represent a 3D convolutional neural network which is an extension of 2D ConvNet for video domain, and applied it to a large-scale video dataset. However, they could not achieve the state-of-the-arts results of classic methods with hand-crafted features. Simonyan~\etal~\cite{zisserman14} designed a two-stream architecture of ConvNets to capture appearance information from individual frames as well as motion information between consecutive frames; \textbf{\em1) Spatial stream} captures appearance information from any individual frame using a pre-trained CNN based on ImageNet dataset. \textbf{\em2) Temporal stream} extracts optical flow to capture motion along the video frames. In very recent work, Wang~\etal~\cite{wang0T15} presented a video representation model called trajectory-pooled deep-convolutional descriptor (TTD) and showed superior results on action recognition task. They used both hand-crafted features to extract trajectories and a two-stream ConvNet for extracting feature maps. TTD is extracted through sampling and pooling trajectories over the convolutional feature maps.\\
\noindent\textbf{Hierarchal structure} models are becoming popular in action recognition methods because of their ability to represent the information of a video in a multi-level fashion. This is beneficial for describing complex human actions~\cite{zamir15, lan2014, tang12}. Lan\etal~\cite{zamir15} present an unsupervised method for building hierarchical models to represent human actions using multiple finer-grained elements, called Mid-level Action Elements ({\em MAE}s). Each {\em MAE} describes an individual action-related segments in a video. The method can automatically discover action-related segments to build a hierarchy of {\em MAE}s for the given action. 
 
 We build a hierarchical model inspired by the Felzenszwalb~\etal~\cite{shapetree} method for representing shapes in a hierarchy of curves. Instead of {\em MAE}s we introduce sub-actions, where each sub-action represents a different level of granularity of the entire action.\\

 Our contributions is a novel method for human actions representation in video, using CNN features, where the underlying network is trained only on images. We parse a complex action into a set of simple sub-actions through a hierarchical structure. Additionally, we introduce a method based on binary coding to detect sub-actions automatically. Moreover, this model represents a video compactly.

\section{CNN Feature Pyramid Architecture}\label{secpyramid}
In this section we describe our proposed method which captures human motion through a hierarchical structure.
The pipeline of our proposed method is shown in Figure~\ref{fig:overview}. It includes four phases (1) Spatial and temporal feature extraction (2) Building a pyramid (3) Creating a video representation (4) Classification. In the following sections we describe each step in detail.

\subsection{Spatial and Temporal}\label{seccnn}
  \noindent\textbf{Spatial Features:} We simply use the AlexNet architecture \cite{alexnet} pre-trained on ImageNet dataset. We utilize the output of $fc7$ layer of this CNN for representing spatial information. \\
  \noindent\textbf{Temporal Component:} More important is how we use CNN features to represent temporal variations in a video.
 While previous CNN based methods for action recognition  use optical flow as a second channel to capture motion information, instead we use what we refer to as "CNN flow". We will see that CNN flow captures informative features about image movement. Figure \ref{fig:pyramid} shows samples of CNN flow computed from a golf pitching snippets.
 
 Given a snippet of a video, we next describe a hierarchical model that computes the flows in the CNN feature space (CNN flow) in a coarse to fine fashion. This idea was first used by \cite{shapetree} to describe curves and shapes in a coarse to fine fashion. We then compute a semantic feature for each video snippet. Later, in section \ref{secsnippet} we explain how we select snippets from a video. We consider each snippet as a temporal word. In order to create a vector representation of an entire video, we adopt a bag-of-words technique to construct a histogram of temporal words.
 
\subsection{Pyramid of CNN Flows}

Most actions consist of sub-actions. For example answering phone includes sub-actions such as stretching arm, grabbing the phone, pulling the arm. Hence, to recognize an entire action, we should recognize these sub-actions.

\begin{figure*}
\begin{center}
\includegraphics[width=\textwidth]{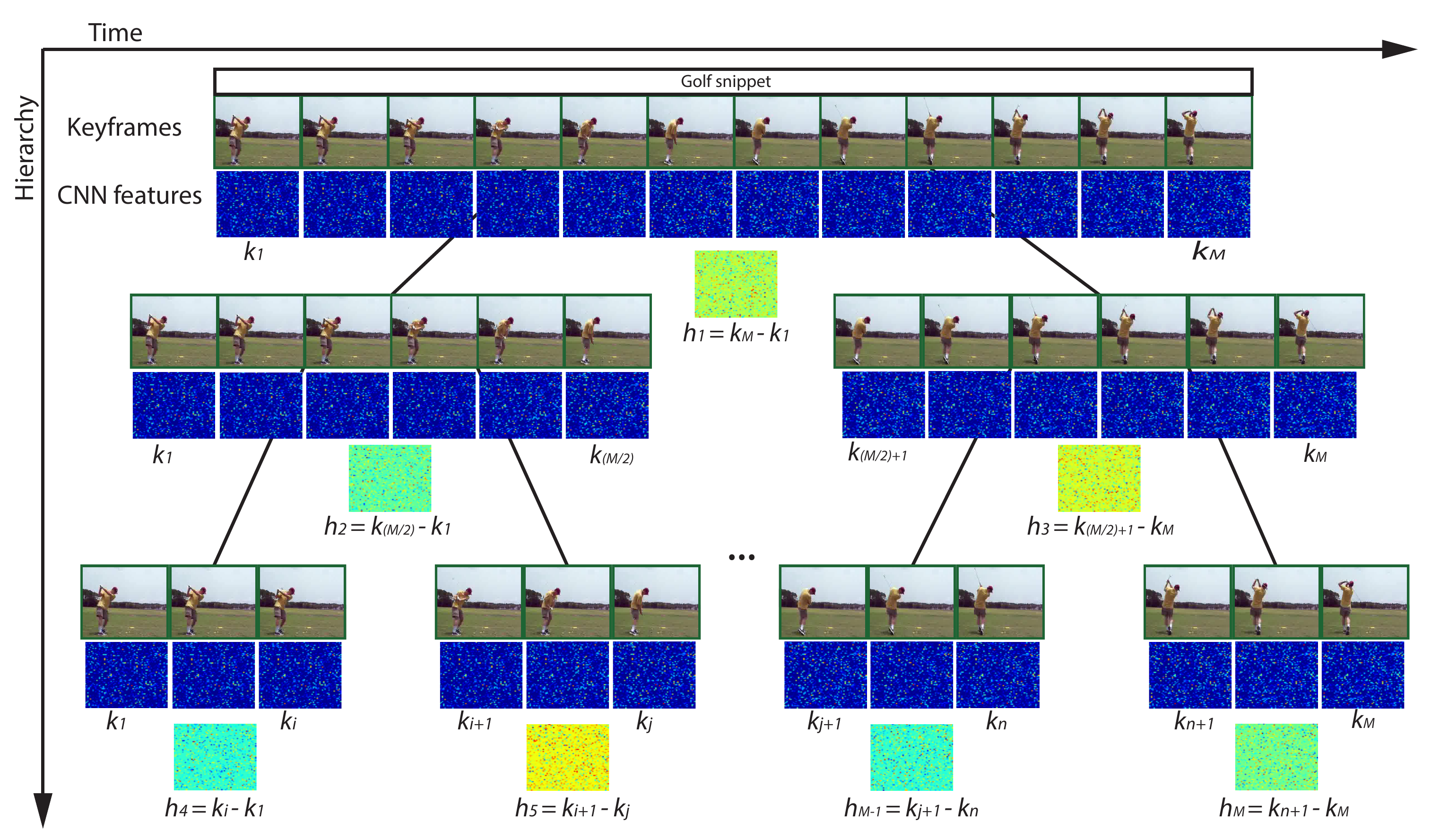}
\end{center}
\caption{Illustration of Hierarchical model for a sample snippet: all frames of given snippet placed on the first level of the pyramid. Snippet frames represented by correspond deep-learned feature vectors $K_{1..M}$. output of this level is a feature vector calculated from the difference between first and last key-frame in the snippet denoted by $h_{1}$. In the second level snippet breaks into two equal parts and for each part we calculate a feature vector from the difference between last and first key-frame in that part denoted by $h_{2}$ and $h_{3}$. This procedure continue till the last level in the pyramid. Finally, stacking all calculated feature vectors $h$ and deliver high-dimensional feature vector as pyramid output to represent the snippet.}
\label{fig:pyramid}
\end{figure*}

 We describe a method which can extract expressive and discriminative sub-actions from a video snippet.
In a given video snippet, inspired from \cite{shapetree}, we capture coarse information by looking at far apart frames in the snippet and compute the CNN flows between those frames. At the next level of the hierarchy, we cut the snippet in half and compute the CNN flow for each sub-snippet. This process creates a binary tree structure; each node in the tree is a CNN flow for a sub-snippet of the video. The higher levels of the binary tree (closer to the root) represent the coarse information of the motion in the snippet and the lower levels (closer to the leaves) represent fine motion information. Figure \ref{fig:pyramid} depicts a schematic illustration of our hierarchical structure for modeling the motion information in a video snippet. 

 A vector representation can be obtained by stacking all the CNN flows across the binary tree. This vector representation may be very high dimensional because each CNN feature has 4096 dimension. Therefore, we reduce the dimension by applying PCA at each level of the tree.

Formally, let \textit{$V_{n}$}, be a video snippet with \textit{$V_{n}=\{v_{i} : i= 1,..., M_{n} \}$}. We extract CNN feature vectors \textit{$X_{n}$} from the frames, which is denoted by \textit{$X_{n}=\{x_{i} : i= 1,..., M_{n} \}$}. Afterward, we take  frames \textit{$K_{n}=\{k_{i} : i= 1,..., M_{m} \}$} , where \textit{$k_{i} (i= 1,..., M_{m})$} is the CNN feature vector of frame $i$. In the next step we organize the keyframes into a hierarchy and denote each level of computed features in the hierarchy as \textit{$H_{n}=\{h_{i} : i= 1,..., M_{n} \}$}, where \textit{$h_{i}$} is the CNN flow; the difference between last keyframe and first keyframe in that particular pyramid level.  The first level of the pyramid computes a feature vector from the difference between last keyframe and first keyframe in the video snippet:
\[h_{1} = k_{M} - k_{1}\]
In the next level of pyramid the video snippet will be divided in two parts and computes \textit{$h_{2}$} and \textit{$h_{3}$} recursively;
\[h_{2} = k_{(M/2)} -  k_{1}\] and , \[h_{3} = k_{M} -  k_{(M/2)+1}\]
We follow the same procedure in all the levels of pyramid, and the final feature vector would be:
\[C_{h_{i}}= [h_{1}, h_{2}, ..., h_{M}] \] 
in any individual level of pyramid the length of feature is 4096 (same as CNN features). To reduce the dimension, we applied PCA on each level of the pyramid to decrease the dimensions.


\subsection{Snippet Selection}\label{secsnippet}

To capture motion across the video, we need to select parts of video which are discriminative and include temporal information to describe a particular motion. CNN flow features in a high frame rate video may be so tiny that does not provide any information. Therefor, selecting the informative parts of video is critical. We need to select a subset of frames, which can represent a summarization of the video. We refer to this subset of frames as key-frames. Here, we discuss about two different approaches to find key-frames. We consider frames between two consecutive key-frames as a snippet. In following section we represent two snippet selection strategies based on key-frames.

To select snippets, we consider two approaches; overlapping windows and keyframe selection using binary codings of the features.\\

\noindent\textbf{Overlapping Windows:} Given a video, we sample the key-frames by skipping a fixed number of frames. We choose a window of length $l$ then we overlaps the windows by stride of size $s$. It means we move the window and jump over each $s$ frames. We consider each window as a snippet. We varies the $l$ and $s$ and generates several video snippets.\\

\begin{figure}
\small{(a) input video}
\begin{center}
\includegraphics[scale=0.35]{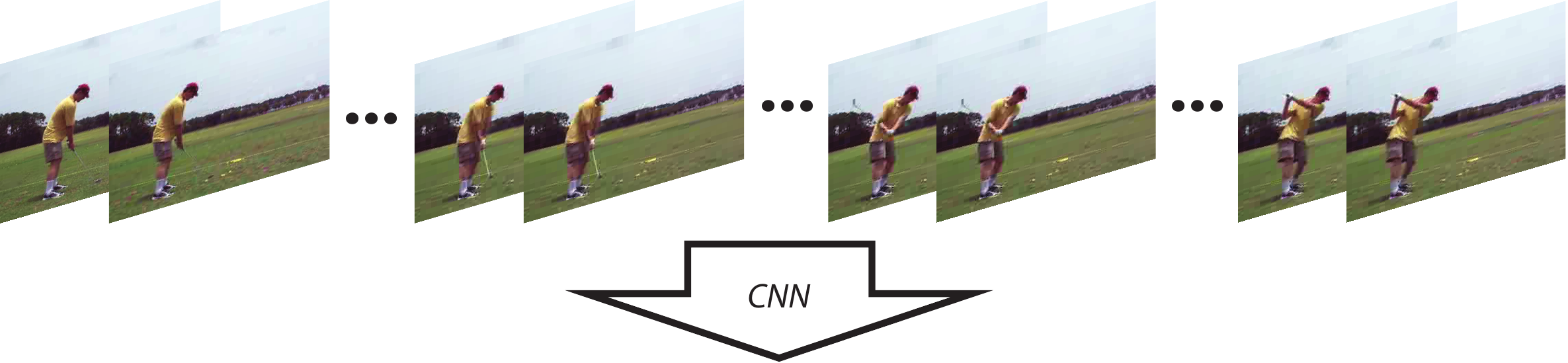}
\end{center}
\small{(a) extract CNN features}
\begin{center}
\includegraphics[scale=0.35]{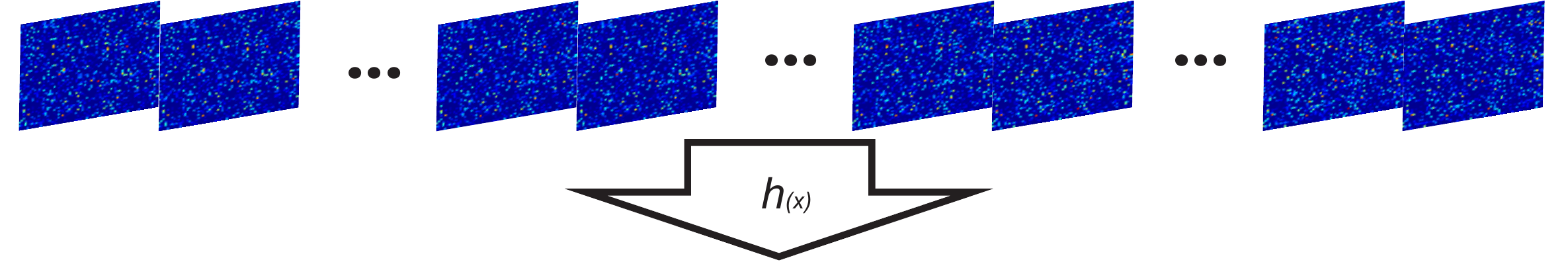}
\end{center}
\small{(c) extract binary codes}
\begin{center}
\includegraphics[scale=0.35]{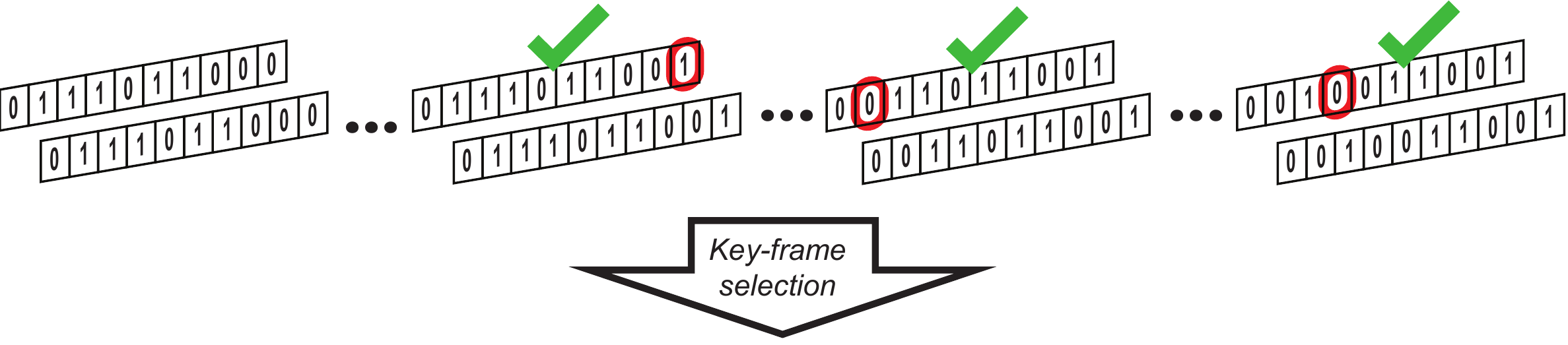}
\end{center}
\small{(d) key-frame selection}
\begin{center}
\includegraphics[scale=0.35]{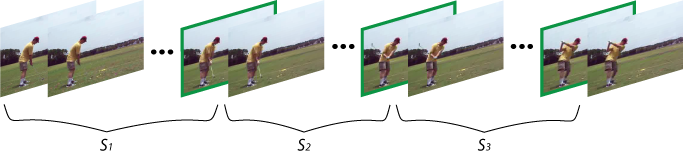}
\end{center}
\caption{Extract Binary key-frames; (a) First parse input video to frames, (b) extract deep-learned features for video frames, (c) then make binary vector for extracted feature vector. Afterward, we observe changing of hamming distances between two consecutive frames during the time. Once any change happened in binary codes between two frames we select the frame as key-frame, denoted by a green mark. (d) sequence of frames between two consecutive key-frames considered as a snippet (noted by $S_{1},~S_{2},~S_{3}$)}
\label{fig:indexing}
\end{figure}

\noindent\textbf{Key-frame Coding:} 
The Overlapping Windows is a kind of brute forcing strategy. Its not efficient and neither optimized. Instead, we propose an efficient technique that can generates expressive video snippets. We expect a video snippet to have enough motion information that can express a part of action and at the same time it should not be so long that includes irrelevant motions. We create a binary code of length $b$ for each frame on CNN feature space. These binary codes should have similar hamming distance for similar points in the CNN feature space. We use Locality Sensitive Hashing (LSH) to hash each frame of a video into a binary space. By moving across the frames, we pick the frames that their binary codes are different from their previous frames. We consider those frames as key-frames and a snippet will be a part of video between two consecutive key-frames. By varying the code length $b$ we can obtain longer or shorter snippets. The longer the code length, the shorter the snippet size. Figure \ref{fig:indexing} illustrates our approach for extracting the video snippets based on key-frame coding.     

Difference between features in two consequent frames, in a high frame rate video, is negligible; otherwise, there is a magnificent movement through the frames. At this moment, the binary values in vector emerge to change and potentially it shows semantic changes. 


\subsection{Bag of snippets} 

To evaluate the Pyramid CNN, we use a standard bag-of-features approach. We first construct a codebook for each level of the pyramid. We fix the number of dictionary elements to 4000 for each level. To limit the complexity, we initialize k-means 10 times and keep the result with the lowest error on validation set.

 Descriptors are assigned to their closest dictionary elements using Euclidean distance. The resulting histograms of temporal word(snippet) occurrences are used as video descriptors. For classification we use a non-linear SVM with a ${\chi}^2$-kernel. In the case of multi-class classification, we use a one-against-rest approach and select the class with the highest score.

\section{Experiments}\label{secexperiment}
In order to compare our method with state-of-the-arts methods, we perform experiments on three action datasets. 

\begin{table*}
\begin{center}
				\begin{tabular}{|c|c|c|c|c|c|}
			\hline
			\multicolumn{2}{|c|}{\textbf{KTH}}&\multicolumn{2}{|c|}{\textbf{UCF Sport}}&\multicolumn{2}{|c|}{\textbf{UCF-11 Human Action}} \\
			\hline
			\hline
			Method 							  & EER $\;\;$ & Method 							& EER $\;\;$ & Method 						& EER $\;\;$ \\
			\hline
			Laptev \etal~\cite{hoha}  				   & 91.8\%  & Souly \& Shah~\cite{shah15}   			& 85.1\%	& 								& \\
			Yuan \etal~\cite{subvolume} 			   & 93.7\%  & Wang \etal~\cite{wang09evaluationof}  	& 85.6\% 	& Incremental Activity Modeling~\cite{incrimental} & 54.5\% \\
			Le \etal~\cite{leng11} 						   & 93.9\%  & Le \etal~\cite{leng11} 						& 86.5\% 	& Liu \etal~\cite{ucf11} 				& 71.2\% \\
			Gilbert \etal~\cite{gilbert11} 			   & 93.9\%  & Kovashka \& Grauman~\cite{kovashka10} 	& 87.2\% 	& Ikizler-Cinbis \etal~\cite{IkizlerCinbis} 	& 75.2\% \\
			Dense Trajectory~\cite{densetraj} 		   & 94.2\% & Dense Trajectory~\cite{densetraj} 		& 89.1\% 	& Dense Trajectory~\cite{densetraj} 		& 84.2\% \\
			Kovashka \& Grauman~\cite{kovashka10}    & 94.5\% & Weinzaepfel~\etal~\cite{cordelia15} 		& 90.5\% 	& Jungchan Cho \etal~\cite{cho14}		 & 88.0\% \\
			\hline
			\hline
			Baseline proposed 					   & 74.5\% & Baseline proposed 					& 88.1\% 	& Baseline proposed 				& 77.1\% \\
			Snippet proposed 					   & 94.1\% & Snippet proposed 					& \textbf{97.8}\% & Snippet proposed 			& \textbf{89.5}\% \\
			Binary proposed 			        & \textbf{95.6}\% & Binary proposed 						& 94.8\% 	& Binary proposed 					& 84.3\% \\
			\hline
		\end{tabular}
	\end{center}

	\caption{Comparison of our results to the state-of-the-arts on action recognition datasets KTH, UCF Sport and UCF-11. Our results are listed in three parts, the proposed baseline according appearance-based features, Overlapping window approach, and binary-based snipped selection technique.}
	\label{results}
\end{table*}


\begin{table}
	\begin{center}
		\begin{tabular}{|c|c|c|c|}
			\hline
			\multicolumn{4}{|c|}{\textbf{UCF Sport}} \\
			\hline
			\hline
			Action Class & Binary & Snippet & Baseline \\
			\hline
			diving 		 & 100\% 	& 100\% 		& 100\% \\
			golf-swing 	& 86\% 	&  94\% 		& 86\% \\
			kicking 		& 98\% 	& 100\% 	    & 85\% \\
			lifting 		& 100\% 	& 100\%	    & 100\% \\
			horse-riding 	& 94\% 	& 100\%	    & 86\% \\
			running 		& 98\%	& 96\%	    & 59\% \\
			skateboarding 	& 86\%	& 92\%	    & 93\% \\
			bench-swing 	& 98\%	& 100\%	    & 95\% \\
			swinging 		& 100\%	& 100\%	    & 100\% \\
			walking 		& 88\%	& 96\%	    & 76\% \\
			\hline
		\end{tabular}
	\end{center}
	\caption{Accuracy  per action class in the UCF Sport dataset.}
	\label{tbl:classsport}
\end{table}
\begin{table}
	\begin{center}
		\begin{tabular}{|c|c|c|c|}
			\hline
			\multicolumn{4}{|c|}{\textbf{KTH Dataset}} \\
			\hline
			\hline
			Action Class & Binary & Snippet & Baseline \\
			\hline
			boxing 		    	& 100\% 	& 100\% 	& 97.2\% \\
			handclapping 		& 95.4\% 	& 91.7\% 	& 91.6\% \\
			handwaving 	    	& 97.3\% 	& 94.5\% 	& 80.5\% \\
			jogging 			& 94.8\% 	& 94.5\%	& 50.0\% \\
			running	 	    	& 86.2\% 	& 83.9\%	& 44.4\% \\
			walking 			& 100\%	& 100\%	& 83.3\% \\
			\hline
		\end{tabular}
	\end{center}
	\caption{Accuracy per action class in the KTH dataset.}
	\label{tbl:classkth}
\end{table}

 \begin{figure}
\subfigure[KTH dataset] {
\includegraphics[scale=0.5]{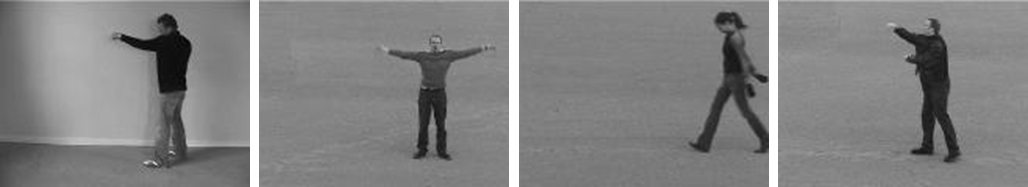}
}
    \subfigure[UCF 11 action dataset] {
\includegraphics[scale=0.5]{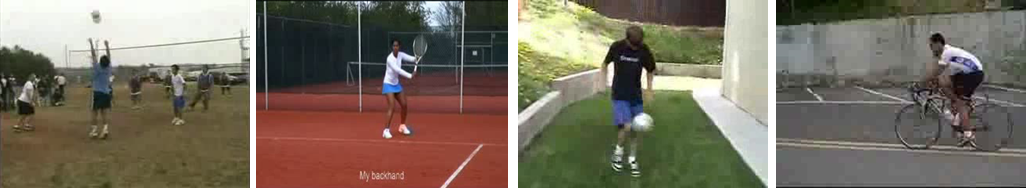}
    }
     \subfigure[UCF sport] {
\includegraphics[scale=0.5]{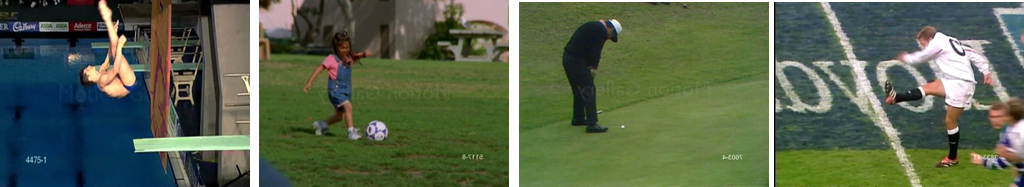}
    }
    \caption{Samples of experimental action datasets}
    \label{fig:db}    
\end{figure}
\subsection {Datasets}\label{dataset}

We evaluate our method on three different datasets, including KTH~\cite{kth}, UCF-11~\cite{ucf11} and UCF sport~\cite{ucfsport}. Samples are shown in Figure~\ref{fig:db}.\\
\noindent\textbf {KTH Dataset}~\cite{kth} includes six action classes: walking, jogging, running, boxing, hand waving and hand clapping. Each action class includes 25 subjects which perform the actions several times in four different scenarios (indoors, outdoors, outdoors with scale variation, outdoors with different clothes). The database contains 600 video sequences, which most of sequences are captured in homogeneous and static background. Our experimental setups are the same as previous works~\cite{kth}: \eg sequences are divided into two sets, test and train. Test set includes all the sequences from 9 subjects indicated by 2, 3, 5, 6, 7, 8, 9, 10, and 22. The rest of the samples from other subjects (16 subjects)are inside the training set (the remaining 16 subjects). following~\cite{kth}, we train a multi-class classifier and report the average of accuracy over all classes as performance measure.\\
\noindent\textbf{UCF-11 Human Action Dataset}~\cite{ucf11} is more challenging dataset in term of large variations in camera motion, illumination, viewpoint, cluttered background, \etc . UCF-11 is a collection of 11 human action categories, 1600 sequences in total. Action categories include: basketball shooting, biking/cycling, diving, golf swinging, horse back riding, soccer juggling, swinging, tennis swinging, trampoline jumping, volleyball spiking, and walking with a dog. We should note that UCF-11 is the extended version of UCF Youtube action dataset, which was a collection of 1200 video samples divided in these 11 action categories. Similar to the original setup~\cite{ucf11} we use leave one out cross validation (LOOCV) for a per-defined set of 25 folds. Performance measure is calculated by average accuracy over all classes.\\
\noindent\textbf{UCF Sport Dataset}~\cite{ucfsport} contains 10 sport actions in 150 video samples: swinging, diving, kicking ball, weight lifting, horse riding, running, skateboarding, swinging (on the bench), golf swinging and walking. We used the setup recommended in~\cite{ucfsport}, using a Leave one out cross validation (LOOCV) scheme, train a multi-class classifier and report the average accuracy over all iterations.

\subsection {Experimental Setting}

Evaluation procedure is designed in two experimental cases. In regards to two different snippet extraction methods, the first experimental case is based on overlapping windows and the second experiment performs under binary-based snippet extraction. In following, we describe the setup of each experimental case as well as our baseline in details.\\
\noindent\textbf {Baselines}
The baseline of our experimental cases is designed to follow two purposes; first, evaluate our method disregards to the state-of-the-arts, second, discover the fact that our method actually capture the motion information regardless to visual appearance and background. Hence, we consider only visual information for feature extraction. We consider fixed-sized snippets (in our experiments 20-frames snippet windows), select alternate frames in each snippet and stack correspond deep-features to build the snippet descriptors. Then, we apply PCA for feature reduction. This technique leads to generate only visual representation of videos. The rest, would be similar to our standard experimental pipeline setup.\\
\noindent\textbf {Experimental Cases}
As mentioned above, we consider two different experimental cases for the snippet extraction methods:\\
\textbf {\em1) overlapping windows} the first experimental case is the snippet-based, which has static pre-defined snippet size. The first and last frame of any snippet indicated as key-frame. We applied several experiments to find most appropriate snippet size. According to the results, the best performance is achieved on snippet length by 20-frames. Also, considering N-frames overlapping on snippets.\\ \textbf{\em2) binary key-frame selection snippet} the second experimental case is based on detected key-frames by binary method. In the snippet-based technique, each snippet comes with a different size. The size of binary code is fixed on 16-bits. The criteria to select a frame as key-frame is to have a hamming distance equal to $1$ between two consecutive frames.
 We employed {\em Iterative Quantization} (ITQ)~\cite{itq11} hashing method, which is a method based on LSH to learn binary attributes during several iterations. The results of this binary code technique could outperform other state-of-the-arts hashing methods. We applied ITQ under 50 iterations to compute binary codes of any individual extracted feature vector with original length of 4096.\\
\noindent\textbf{Feature Extraction} 
Most existing works in action recognition use complex structure of handcrafted features as inputs, instead we focused on CNN global features, which has shown promising performance on recognition tasks recently\cite{alexnet}. More precisely, in our implementation, we choose the output of $fc7$ layer of AlexNet layers \cite{alexnet}. More precisely, to extract AlexNet features, we use CAFFE \cite{cafe14}, due to their promising performance and accuracy. The Outputs of Deep-net represent spatial features which capture static visual appearance information, obtained from single frame images (resized to $224 \times 224 \times 3$).\\
\noindent\textbf {Pyramid Setup}
The pyramid aims to describe the dynamic motion information using a hierarchical structure. In our setup the pyramid consist 4 levels. First, video parse to snippets, then individual snippet frames feed to the first level of pyramid. In the second level of pyramid, snippet frames divided by two parts, then in the third level snippet takes in four parts, and so on.  In the last level the snippet is divided into 10 parts, to capture finer sub-actions.\\
After computing full feature vector across the pyramid levels, the snippet is represented as a single feature vector by length of 69631 ($17 \times 4096$). In order to reduce the dimension of the final feature vector a PCA applied on each pyramid level separately. Computed feature size on each level is 4096 which is reduced to 100 after applying PCA. Thus, the final feature vector has a dimension of 1700, which represents entire snippet.


\subsection {Experimental Results}

We applied our method on three datasets: KTH, UCF-Sport Action and UCF-11 Action datasets. We summarized the results of action recognition accuracy across these datasets in Table~\ref{results}. The table shows the results of our approach in compare with the state-of-the-arts. Our results improves the accuracy of action recognition in these datasets.

\begin{figure}
	\begin{center}

	\centering
	\scalebox{0.65}{
	\begin{tabular}{c | c c c c c c c c c c c c}
		\multicolumn{1}{c}{} & & \multicolumn{9}{c}{Prediction}& \\ \cline{3-13}
		 \multicolumn{1}{c}{} & & \rotatebox[origin=c]{90}{shoot} & \rotatebox[origin=c]{90}{biking} & \rotatebox[origin=c]{90}{diving} & \rotatebox[origin=c]{90}{g-swing} & \rotatebox[origin=c]{90}{h-riding} & \rotatebox[origin=c]{90}{juggling} & \rotatebox[origin=c]{90}{swing} & \rotatebox[origin=c]{90}{tennis}& \rotatebox[origin=c]{90}{jump}& \rotatebox[origin=c]{90}{v-spiking}& \multicolumn{1}{c}{\rotatebox[origin=c]{90}{walking}}\\
		\multirow{11}{*}{\rotatebox[origin=c]{90}{Truth}}
		
		 &shoot	& \cw{100}& \cb{0.0}	& \cb{0.0}	& \cb{0.0} 	& \cb{0.0} 	& \cb{0.0}	& \cb{0.0} 	& \cb{0.0}	& \cb{0.0}	& \cb{0.0}	& \cb{0.0} \\ 
		 &biking	& \cb{0.0}	& \cw{94} & \cb{2.0}& \cb{0.0}	& \cb{2.0} 	& \cb{2.0}	& \cb{0.0} 	& \cb{0.0}	& \cb{0.0}	& \cb{0.0}	& \cb{0.0} \\ 
		 &diving	& \cb{0.0}& \cb{0.0}	& \cw{100} & \cb{0.0}& \cb{0.0} 	& \cb{0.0}	& \cb{0.0} 	& \cb{0.0}	& \cb{0.0}	& \cb{0.0}	& \cb{0.0} \\ 
		 &g-swing	& \cb{0.0}	& \cb{0.0}	& \cb{0.0}	& \cw{100}& \cb{0.0} & \cb{0.0}	& \cb{0.0} 	& \cb{0.0}	& \cb{0.0}	& \cb{0.0}	& \cb{0.0} \\ 
		 &h-riding	& \cb{0.0}	& \cb{0.0}	& \cb{.0}	& \cb{0.0}	& \cw{100} & \cb{0.0}& \cb{0.0} & \cb{0.0}	& \cb{0.0}	& \cb{0.0}	& \cb{0.0} \\ 
		 &juggling	& \cb{0.0}	& \cb{0.0}	& \cb{0.0}	& \cb{0.0}	& \cb{0.0} 	& \cw{96}& \cb{0.0} & \cb{0.0}	& \cb{0.0}	& \cb{0.0}	& \cb{0.0} \\ 
		 &swing	& \cb{0.0}	& \cb{0.0}	& \cb{0.0}	& \cb{0.0}	& \cb{0.0} 	& \cb{0.0}	& \cw{96} & \cb{0.0}& \cb{0.0}	& \cb{0.0}	& \cb{4.0} \\ 
		 &tennis	& \cb{0.0}	& \cb{0.0}	& \cb{0.0}	& \cb{0.0}	& \cb{0.0} 	& \cb{0.0}	& \cb{0.0} 	& \cw{92}& \cb{0.0}& \cb{0.0}	& \cb{4.0} \\ 
		 &jump	& \cb{0.0}	& \cb{0.0}	& \cb{0.0}	& \cb{0.0}	& \cb{0.0} 	& \cb{0.0}	& \cb{0.0} 	& \cb{0.0}	& \cw{100}& \cb{0.0}& \cb{0.0} \\ 
		 &v-spiking& \cb{0.0}	& \cb{0.0}	& \cb{0.0}	& \cb{0.0}	& \cb{0.0} 	& \cb{0.0}	& \cb{0.0} 	& \cb{0.0}	& \cb{0.0}	& \cw{100}& \cb{0.0} \\ 
		 &walking	& \cb{0.0}	& \cb{2.0}	& \cb{0.0}& \cb{0.0}	& \cb{0.0} 	& \cb{0.0}	& \cb{0.0} 	& \cb{2.0}	& \cb{0.0}	& \cb{0.0}	& \cw{96} \\
	\end{tabular}}	
	\end{center}
	\caption{Confusion matrix per action class for UCF-11 action dataset.}
	\label{fig:fconfucf11}
\end{figure}
\begin{figure}
	\begin{center}

	\scalebox{1}{
	\begin{tabular}{c | c c c c c c c c c c c c}
		\multicolumn{1}{c}{} & & \multicolumn{5}{c}{Prediction}& \\ \cline{3-8}
		 \multicolumn{1}{c}{} & & \rotatebox[origin=c]{90}{box} & \rotatebox[origin=c]{90}{h-clp} & \rotatebox[origin=c]{90}{h-wav} & \rotatebox[origin=c]{90}{jog} & \rotatebox[origin=c]{90}{run} & \rotatebox[origin=c]{90}{walk}\\
		\multirow{6}{*}{\rotatebox[origin=c]{90}{Truth}}
		
		 &boxing	& \cw{100}  & \cb{0.0}	& \cb{0.0}	& \cb{0.0} 	& \cb{0.0} 	& \cb{0.0}	\\ 
		 &h-clapp	& \cb{4.6}	& \cw{95.4} & \cb{0.0}  & \cb{0.0}	& \cb{0.0} 	& \cb{0.0}	\\ 
		 &h-wav	    & \cb{0.0}  & \cb{2.7}	& \cw{97.3} & \cb{0.0}  & \cb{0.0}  & \cb{0.0}	 \\ 
		 &jogging	& \cb{0.0}	& \cb{0.0}	& \cb{0.0}	& \cw{94.8} & \cb{2.5}  & \cb{2.7}	\\ 
		 &running	& \cb{0.0}	& \cb{0.0}	& \cb{0.0}	& \cb{11.1} & \cw{86.2} & \cb{2.7}\\ 
		 &walking	& \cb{0.0}	& \cb{0.0}	& \cb{0.0}	& \cb{0.0}	& \cb{0.0} 	& \cw{100} \\ 		 
	\end{tabular}}	
	\end{center}
	\caption{Confusion matrix per action class for KTH dataset based on binary key-frame selection snippet.}
	\label{fig:confkth}
\end{figure}

\begin{figure*}
\centering
\subfigure[Change binary size] {
\begin{tikzpicture}[scale=0.66]
    \begin{axis}[
    	grid=major,
        xlabel=$binary~size$,
        ylabel=$accuracy \%$]
    \addplot[smooth,mark=*,blue] plot coordinates {
        (8,88.8)
        (10,92.2)
        (16,95.6)
        (20,94.3)
        (32,96.3)
    };
    \end{axis}
    \end{tikzpicture}
	}
    \subfigure[Change overlapping window size] {
    \begin{tikzpicture}[scale=0.66]
    \begin{axis}[
    	grid=major,
        xlabel=$window~size$,
        ylabel=$accuracy$,
        xtick={20,30,40,50}]
    \addplot[smooth,mark=*,blue] plot coordinates {
        (20,94.1)
        (30,92.7)
        (40,93.9)
        (50,90.5)
    };
    \end{axis}
    \end{tikzpicture}
    }
     \subfigure[Change pyramid levels] {
    \begin{tikzpicture}[scale=0.66]
    \begin{axis}[
    	grid=major,
        xlabel=$pyramid~level$,
        ylabel=$accuracy$,
        xtick={1,2,3,4},
        xmin=0]
    \addplot[smooth,mark=*,blue] plot coordinates {
        (1,89.0)
        (2,93.9)
        (3,92.0)
        (4,94.1)
    };
    \end{axis}
    \end{tikzpicture}
    }
 \caption{Experimental results for KTH dataset; recognition accuracy observation under parameters: (a) change in size of binary code, (b) different overlapping window size (c) various pyramid levels}
    \label{fig:expkth}    
\end{figure*}
    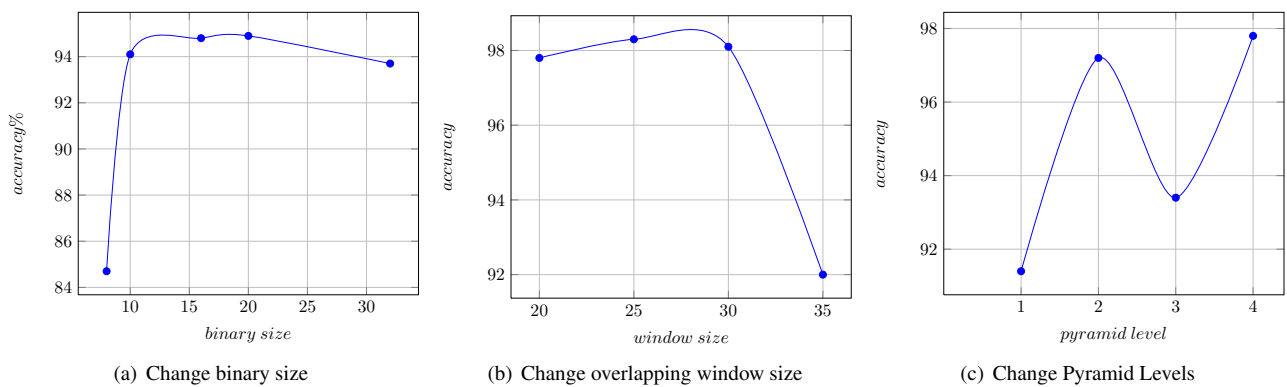
\begin{figure*}
\centering
\subfigure[Change binary size] {
\begin{tikzpicture}[scale=0.66]
    \begin{axis}[
    	grid=major,
        xlabel=$binary~size$,
        ylabel=$accuracy \%$]
    \addplot[smooth,mark=*,blue] plot coordinates {
        (8,84.7)
        (10,94.1)
        (16,94.8)
        (20,94.9)
        (32,93.7)
    };
    \end{axis}
    \end{tikzpicture}
	}
    \subfigure[Change overlapping window size] {
    \begin{tikzpicture}[scale=0.66]
    \begin{axis}[
    	grid=major,
        xlabel=$window~size$,
        ylabel=$accuracy$]
    \addplot[smooth,mark=*,blue] plot coordinates {
        (20,97.8)
        (25,98.3)
        (30,98.1)
        (35,92)
    };
    \end{axis}
    \end{tikzpicture}
    }
     \subfigure[Change Pyramid Levels] {
    \begin{tikzpicture}[scale=0.66]
    \begin{axis}[
    	grid=major,
        xlabel=$pyramid~level$,
        ylabel=$accuracy$,
        xtick={1,2,3,4},
        xmin=0]
    \addplot[smooth,mark=*,blue] plot coordinates {
        (1,91.4)
        (2,97.2)
        (3,93.4)
        (4,97.8)
    };
    \end{axis}
    \end{tikzpicture}
    }
    \caption{Experimental results for UCF sport dataset; recognition accuracy observation under parameters: (a) change in size of binary code, (b) different overlapping window size (c) various pyramid levels}
    \label{fig:expsport}    
\end{figure*}
\subsubsection {Results on KTH Dataset}\label{seckthexp}
 To evaluate our method on KTH dataset we followed similar setup as~\cite{kth}. Furthermore, we changed several parameters to understand how changing different parameters effects on the final performance. The experimental parameters are listed as below:\\
\noindent\textbf {Binary Size} varies in range of 8, 10, 16, 20, and 32.\\
\noindent\textbf {Overlapping window size} parameter also can effect on the recognition accuracy. We tried 20, 30, 40, and 50 frame length for snippets.\\
\noindent\textbf {Pyramid levels} indicate how much the fine sub-actions can be captured. The longer pyramid captures smaller sub-actions. In our experiments we tried pyramid with 1, 2, 3, and 4 levels.\\
The results under different parameters is shown in Figure~\ref{fig:expkth}. Figure explains that increasing the length of overlapping window decrease the recognition accuracy.  More levels in the hierarchy can lead to better performance, since model would be able to capture more detailed information of the motion in the snippet.\\
Figure~\ref{fig:confkth} shows confusion matrix per action class based on on binary key-frame selection snippet method in KTH dataset. The best accuracy appears in binary size 16 with 4 levels in pyramid. A detailed results per action class are shown in Table~\ref{tbl:classkth}; overlapping window snipped, binary keyframe selection, and our baseline.

\subsubsection {Results on UCF Sport Dataset}

  In~\ref{seckthexp}, we present an ablation study of the effect of changing our parameters on final recognition accuracy.  Similar experimental results are presented in Figure~\ref{fig:expsport} for UCF-Sport dataset. Since UCF Sport video samples are short, in this experiments we tried different range for overlapping window size (20, 25, 30, and 35 frame length for snippets). Overall, the best accuracy achieved on the snippet with length 20 and 4 levels in the pyramid.
  
  Figure~\ref{fig:expsport} (a) shows that increasing size of binary vector can improve recognition accuracy. Figure~\ref{fig:expsport} (b) tracks the changes in overlapping window length. Growing the snippet length comes with lower recognition performance, and Figure~\ref{fig:expsport} (c) represents the effect of different levels in pyramid.

\subsubsection {Results on UCF11}

Experiments for UCF-11 dataset are performed under a 25 fold LOOCV technique as suggested in original setup~\cite{ucf11}. We use predefined fixed parameter for binary size 16, overlapping window length 20, and four levels in the pyramid. The results obtained for our proposed methods; snipped, binary, and baseline summerized in Table~\ref{results}, as well as a confusion matrix with more details shown in Figure~\ref{fig:fconfucf11}.

The results indicate the robustness of our method across different parameters. Our method improve action recognition performance over the best state-of-the-arts results.


\section{Conclusion}
We introduced a feature representation for videos that outperforms state-of-the-arts methods on several datasets for action recognition. Our representation is achieved by a hierarchical structure of CNN features where the underlying network trained by image data. Inspired by optical flow we introduced CNN-flow were it is simply subtraction of two CNN features across time. We proposed a novel technique to find key-frames in a video such that the video snippet inbetween two key-frames caries considerable amount of information. 

 Furthermore, the experimental results show the proposed method achieves better performance on action recognition accuracy compared to other reported state-of-the-arts results as well as standard baselines.

{\small
\bibliographystyle{ieee}
\bibliography{egbib}
}

\end{document}